\documentclass{article}

\usepackage[final]{nips_2018}

\usepackage[utf8]{inputenc} 
\usepackage[T1]{fontenc}    
\usepackage{hyperref}       
\usepackage{url}            
\usepackage{booktabs}       
\usepackage{amsfonts}       
\usepackage{nicefrac}       
\usepackage{microtype}      

\usepackage{times}  
\usepackage{helvet}  
\usepackage{courier}  
\usepackage{url}  
\usepackage{graphicx}  

\usepackage{thm-restate}

\usepackage[usenames]{color}
\usepackage[usenames,dvipsnames]{xcolor}
\usepackage{ifthen}
\usepackage{mathrsfs}
\usepackage{graphicx}

\usepackage{amsmath}
\usepackage{amsthm}
\usepackage{algorithmic}
\usepackage[boxed]{algorithm}
\usepackage{amssymb}

\usepackage{fullpage}
 \usepackage{natbib}
\usepackage{bm}
\usepackage{indentfirst}

\usepackage{hyperref}
\hypersetup{
  colorlinks   = true, 
  urlcolor     = blue, 
  linkcolor    = blue, 
  citecolor   = blue 
}

\usepackage[hyphenbreaks]{breakurl}

\newtheorem{fact}{Fact}
\newtheorem{theorem}{Theorem}

\newtheorem{lemma}{Lemma}

\newtheorem{proposition}{Proposition}

\newcommand{\compilehidecomments}{false}

\ifthenelse{ \equal{\compilehidecomments}{true} }{%
	\newcommand{\wei}[1]{}
	\newcommand{\siwei}[1]{}
}{
	\newcommand{\wei}[1]{{\color{blue!50!black}  [\text{Wei:} #1]}}
	\newcommand{\siwei}[1]{{\color{brown!60!black} [\text{Siwei:} #1]}}
}

\newcommand{\E}{\mathbb{E}}

\newcommand{\B}{\mathbf{B}}

\newcommand{\argmax}{\operatornamewithlimits{argmax}}

\newcommand{\I}{\mathbb{I}}

\allowdisplaybreaks

\title{Multi-armed Bandits with Compensation}

\author{
 Siwei Wang \\
 IIIS, Tsinghua University\\
 \texttt{wangsw15@mails.tsinghua.edu.cn} \\
 \And
 Longbo Huang\\
IIIS, Tsinghua University\\
\texttt{longbohuang@tsinghua.edu.cn}\\
}
 
\begin{document}

\maketitle

\begin{abstract}
We propose and study the known-compensation multi-arm bandit (KCMAB) problem, where a system controller offers a set of arms to many short-term players for $T$ steps. In each step, one short-term player arrives to the system. Upon arrival, the player aims to select an arm with the current best average reward and receives a stochastic reward associated with the arm. In order to incentivize players to explore other arms,  the controller provides a proper payment compensation to players. The objective of the controller is to maximize the total reward collected by players while minimizing the  compensation. 
We first provide a compensation lower bound $\Theta(\sum_i {\Delta_i\log T\over KL_i})$, where $\Delta_i$ and $KL_i$ are the expected reward gap and Kullback-Leibler (KL) divergence between distributions of arm $i$ and the best arm, respectively. %
We then analyze three algorithms to solve the KCMAB problem, and obtain their regrets and compensations. We show that the algorithms all achieve $O(\log T)$ regret and $O(\log T)$ compensation that match the theoretical lower bound. Finally, we present experimental results to demonstrate the performance of the algorithms.  
\end{abstract}

\section{Introduction}

Multi-arm bandit (MAB) is a game that lasts for an unknown time horizon $T$ \citep{Berry1985Bandit,Sutton1998Reinforcement}. In each time slot, the controller has $N$ arms to pull, and pulling different arms will result in  different feedbacks. In the stochastic MAB model \citep{Lai1985Asymptotically},  feedbacks from a single arm follow an associated distribution, which is unknown to the player. These feedbacks are random variables independent of any other events. After pulling the arm, the controller collects a reward that depends on the feedback.  
The controller aims to maximize the sum of rewards during the game by choosing a proper arm to pull in each time slot, and the decision can depend on all available information, i.e., past chosen arms and feedbacks. 
The common metric for evaluating the performance of a policy is the value of regret,  defined as the expected difference between the player's reward and pulling an arm that generates the largest average reward. 

The MAB formulation models the trade-off between exploration and exploitation, where exploration concerns exploring the potential best arms, but can result in pulling sub-optimal arms, while exploitation aims at choosing arms with the current best performance and can lose reward if that arm is in fact sub-optimal. 
Thus, optimizing this trade-off is very important for any controller seeking to minimize regret.  
However, in many real-world applications, arms are not pulled by the controller concerning long-term performance. Instead, actions are taken by short-term players interested in optimizing their instantaneous reward. 
In this case, an important means is to provide monetary compensation to players, so that they act as if they are pulling the arms on behalf of the controller, to jointly minimize regret, e.g.,  \cite{Frazier2014Incentivizing}. Our objective in this paper is to seek an incentivizing policy, so as to minimize regret while not giving away too much compensation. 

As a concrete example,  consider the scenario where a e-commerce website recommends goods to consumers. 
When a consumer chooses to purchase a certain  good, he receives the reward of that good. The website similarly collects the same reward as a recognition of the recommendation.  
In this model, the website acts as a controller that  decides how to provide recommendations. Yet, the actual decision is made by consumers, who  are not interested in exploration and will choose to optimize their reward greedily. 
However, being a long-term player,  the website cares more about maximizing the total reward during the game. As a result, he needs to devise a scheme to influence the choices of  short-term consumers, so that both the consumers and website can maximize their benefits.    
One common way to achieve this goal in practice is that the website offers customized discounts for certain goods to consumers, i.e., by offering compensation to pay for part of the goods. 
In this case, each customer, upon arrival,  will choose the good with largest expected reward plus the compensation. 
The goal of the e-commerce site is to find an optimal compensation policy to minimize his regret, while not spending  too much additional payment. 

It is important to notice the difference between regret and compensation. In particular, regret comes from pulling a sub-optimal arm, while compensation comes from pulling an arm with poor past behavior. 
For example, consider two arms with expected reward $0.9$ and $0.1$. Suppose in the first twenty observations, arm $1$ has an empirical mean $0.1$ but arm $2$ has an empirical mean $0.9$. Then, in the next time slot, pulling arm $2$ will cause regret $0.8$, since its expected gain is $0.8$ less than arm $1$. But in a short-term player's view, arm $2$ behaves better than arm $1$. Thus, pulling arm $2$ does not require any compensation, while pulling arm $1$ needs $0.8$ for compensation. 
As a result, the two measures can behave differently and require different analysis, i.e., regret depends heavily on learning the arms well, while compensation is largely affected by how the reward dynamics behaves.

There is a natural trade-off between regret and compensation. If one does not offer any compensation, the resulting user selection policy is greedy, which will lead to a $\Theta(T)$ regret. 
On the other hand, if one is allowed to have arbitrary compensation, 
one can achieve an $O(\log T)$ regret with many existing algorithms. 
The key challenge in obtaining the best trade-off between regret and compensation lies in  that the compensation value depends on the random history. 
As a consequence, different random history not only leads to different compensation value, but also results in different arm selection. 
Moreover, in practice, the compensation budget may be limited, e.g., a company hopes to maximize its total income which equals to reward subtracts compensation.  
These make it hard to analyze its behavior.

%


%

\subsection{Related works} 
The incentivized learning model has been investigated in prior works, e.g.,  \cite{Frazier2014Incentivizing,mansour2015bayesian,mansour2016bayesian}. In \cite{Frazier2014Incentivizing}, the model contains a prior distribution for each arm's mean reward at the beginning. As time goes on,  observations from each arm update the posterior distributions, and subsequent decisions are made based on posterior distributions. 
The objective is to optimize the total discounted rewards. 
Following their work, \cite{mansour2015bayesian} considered the case when the rewards are not discounted, and they show an algorithm to achieve regret upper bound of $O(\sqrt{T})$. 
In \cite{mansour2016bayesian}, instead of a simple game, there is a complex game  in each time slot that contains more players and actions. 
These incentivization formulations can  model many practical applications, including crowdsourcing and recommendation systems \citep{che2017recommender,papanastasiou2017crowdsourcing}.

In this paper, we focus on the non-Bayesian setting and consider the non-discounted reward. As has been pointed out in \cite{mansour2015bayesian}, the definition of user expectation is different in this case. Specifically, in our setting, each player selects arms based on their empirical means, whereas in the Bayesian setting, it is possible for a player to also consider posterior distributions of arms for decision making. 
%
We propose three algorithms for solving our problem, which adapt ideas from existing policies for stochastic MAB, i.e.,  Upper Confidence Bound (UCB) \citep{Auer2002Finite,gittins1989multi}, Thompson Sampling (TS)  \citep{thompson1933likelihood} and $\varepsilon$-greedy  \citep{Watkins1989Learning}. These algorithms can guarantee $O(\log T)$ regret upper bound, which matches the regret lower bound $\Theta(\log T)$ \citep{Lai1985Asymptotically}. 

Another related bandit model is contextual bandit, where a context is contained in each time slot \citep{Auer2002The,beygelzimer2011contextual,maillard2011adaptive}. 
The context is given before a decision is made, and the reward depends on the context. As a result, arm selection also depends on the given context. 
In incentivized learning, the short-term players can view the compensation as a context, and their decisions are influenced by the context. 
However, different from contextual bandits, where the context is often an exogenous  random variable and the controller focuses on identifying the best arm under given contexts, in our case, the context is given by the controller and itself influenced by player actions. 
Moreover, the controller needs to pay for obtaining  a desired context. What he needs is the best way to construct a context in every time slot, so that the total cost is minimized. 

In the budgeted MAB model, e.g.,  \citep{combes2015bandits,tran2012knapsack,xia2015thompson},  players also need to pay for pulling arms. In this model, pulling each arm costs a certain budget. %
The goal for budgeted MAB is to maximize the total reward subject to the budget constraint. 
The main difference from our work is that in budgeted MAB, 
the cost budget for pulling each arm is pre-determined and it does not change with the reward history. In incentivized learning, however,  different reward sample paths will lead to different costs for pulling the same arm. 

\subsection{Our contributions}
The main contributions of our paper are summarized as follows:
 
\begin{enumerate}
\item  
We propose and study the Known-Compensation MAB problem (KCMAB). In KCMAB, a long-term controller aims to optimize the accumulated reward but has to offer compensation to a set of short-term players for pulling arms. 
Short-term players, on the other hand, arrive at the system and make greedy decisions to maximize their expected reward plus compensation, based on knowledge about previous rewards and offered compensations. 
The objective of the long-term controller is to design a  proper compensation policy, so as to minimize his regret with minimum compensation. 
KCMAB is a non-Bayesian and non-discounted extension of the model in \cite{Frazier2014Incentivizing}. 

\item
In KCMAB, subject to the algorithm having an $o(T^\alpha)$ regret for any $\alpha > 0$, we provide a $\Theta(\log T)$ lower bound for the compensation. 
This compensation lower bound has the same order as the regret lower bound, which means that one cannot expect a compensation to be much less than its regret, if the regret is already small. 

\item 
We propose algorithms to solve the KCMAB problem 
and present their compensation  analysis. Specifically, we provide the analyses of compensation for the UCB policy, a modified $\varepsilon$-greedy policy and a modified-TS policy. All these algorithms have $O(\log T)$ regret upper bounds  while using compensations upper bounded $O(\log T)$, which matches the lower bound (in order). 
 
\item 
We provide experimental results to demonstrate the performance of our algorithms. 
In experiments, we find that modified TS policy behaves better than UCB policy, while the modified $\varepsilon$-greedy policy has regret and compensation slightly larger than those under the modified-TS policy. 
We also compare the classic TS algorithm and our modified-TS policy. The results show that our modification is not only effective in analysis, but also impactful on actual performance. Our results also demonstrate the  trade-off between regret and compensation. 
\end{enumerate}



\section{Model and notations}


In the Known-Compensation Multi-Arm Bandit (KCMAB)  problem,  a central controller has $N$ arms $\{1,\cdots,N\}$. Each arm $i$ has a reward distribution denoted by $D_i$ with support $[0,1]$ and mean $\mu_i$. 
Without loss of generality, we assume $1\ge \mu_1 > \mu_2 \ge \cdots \mu_N \ge 0$ and set $\Delta_i = \mu_1 - \mu_i$ for all $i\ge 2$. The game is played for $T$ time steps. 
In each time slot $t$, a short-term player arrives at the system and needs to choose an arm $a(t)$ to pull. After the player pulls arm $a(t)$,  the player and the controller each receive a reward $X(t)$ from the distribution $D_{a(t)}$, denoted by $X_{a(t)}(t) \sim D_{a(t)}$, which is an independent random variable every time arm $a(t)$ is pulled.  

Different from the classic MAB model, e.g., \cite{Lai1985Asymptotically}, where the only control decision is arm selection, the controller can also choose to offer a compensation  to a player for pulling a particular arm, so as to incentivize the player  to explore an  arm favored by the controller. We denote the paid compensation $c(t) = c_{a(t)}(t)$, and assume that it can depend on all the previous information, i.e., it depends on $\mathcal{F}_{t-1} = \{(a(\tau), X(\tau), c(\tau))| 1 \le \tau \le t-1\}$. Each player is assumed to choose an arm greedily to maximize his total expected income based on past observations. Here income for pulling arm $i$ equals to $\hat{\mu}_i(t) + c_i(t)$, where $\hat{\mu}_i(t) \triangleq M_i(t)/N_i(t)$ is the empirical mean of arm $i$, with  $N_i(t) = \sum_{\tau< t} \I[a(\tau) = i]$ being the total number of times for pulling arm $i$ and $M_i(t) = \sum_{\tau< t} \I[a(\tau) = i]X(t)$ being the total reward collected. 
%
Therefore, each player will pull arm $i = \argmax_j \{\hat{\mu}_j(t) + c_j(t)\}$.

%
%



The  long-term controller, on the other hand, concerns about the expected total reward.  Following the MAB tradition, we define the following total regret: 
\begin{eqnarray*}
Reg(T) = T\max_i \mu_i - Rew(T) = T\mu_1 - Rew(T), 
\end{eqnarray*}
where $Rew(T)$ denotes the expected total reward that the long-term controller can obtain until time horizon $T$. 
We then use $Com_i(T) = \E\left[\sum_{\tau=1}^T \I[a(\tau) = i] c(\tau)\right]$ to denote the expected compensation  paid for arm $i$,  
%
and denote $Com(T) = \sum_i Com_i(T)$ the expected total compensation. 

It is known from \cite{Lai1985Asymptotically}  that $Reg(T)$ has a lower bound of $\Omega(\sum_{i=2}^N {\Delta_i\log T\over KL(D_i,D_1)})$, where $KL(P,Q)$ denotes the Kullback-Leibler (KL)-divergence between distributions $P$ and $Q$, even when a single controller is pulling arms for all time. %
%
Thus, our objective is to minimize the compensation while keeping the regret upper bounded by $O(\sum_{i=2}^N {\Delta_i\log T\over KL(D_i,D_1)})$.

Note that in the non-Bayesian model, if there are no observations for some arm $i$,  players will have no knowledge about its mean reward and they cannot make  decisions. Thus, we assume without loss of generality that in the first $N$ time slots of the game, with some constant compensation, the long-term controller can control the short-term players to choose all the arms once. This assumption does not influence the results in this paper. 

In the following, we will present our algorithms and analysis. Due to space limitation, all complete proofs  in this paper are deferred to  the appendix. We only provide proof sketches in the main text.

\section{Compensation lower bound}

In this section, we first derive a compensation lower bound,  subject to the constraint that the algorithm guarantees an $o(T^\alpha)$ regret for any $\alpha > 0$. 
We will make use of the following simple fact to simplify the computation of compensation at every time slot. 
\begin{fact}\label{Fact_EZ} 
	If the long-term controller wants the short-term player to choose arm $i$ in time slot $t$, then the minimum compensation he needs to pay on pulling arm $i$ is $c_i(t) = \max_j \hat{\mu}_j(t) - \hat{\mu}_i(t)$.
\end{fact}

With Fact \ref{Fact_EZ}, we  only need to consider the case $ c(t) = \max_j \hat{\mu}_j(t) - \hat{\mu}_i(t)$ for   arm $i$.

\begin{theorem}\label{Theorem_LB}
  In KCMAB, if an algorithm guarantees an $o(T^\alpha)$ regret upper bound for any fixed $T$ and any $\alpha > 0$, then there exist examples of Bernoulli Bandits, i.e., arms having reward $0$ or $1$ every time, such that the algorithm must pay $\Omega\left(\sum_{i=2}^N {\Delta_i\log T\over KL(D_i,D_1)}\right)$ for compensation in these examples. 
\end{theorem}

\textbf{Proof Sketch}: Suppose an algorithm achieves an $o(T^\alpha)$ regret upper bound for any $\alpha > 0$. We know that it must pull a sub-optimal arm $i$ for $\Omega\left({\log T\over KL(D_i,D_1)}\right)$ times almost surely \citep{Lai1985Asymptotically}. 
%
Now denote $t_i(k)$ be the time slot (a random variable) that we choose arm $i$ for the $k$-th time. We see that one needs to pay $\E[\max_j \hat{\mu}_j(t_i(k)) - \hat{\mu}_i(t_i(k))] \ge \E[\hat{\mu}_1(t_i(k))] - \E[\hat{\mu}_i(t_i(k))]$ for compensation in that time slot. 
By definition of $t_i(k)$ and the fact that all rewards are independent with each other, we always have $\E[\hat{\mu}_i(t_i(k))] = \mu_i$. 

It remains to bound the value $\E[\hat{\mu}_1(t_i(k))]$. 
Intuitively, when $\mu_1$ is large,  $\E[\hat{\mu}_1(t_i(k))]$ cannot be small, since those random variables are with mean $\mu_1$. Indeed, when $\mu_1 > 0.9$ and $D_1$ is a Bernoulli distribution, one can prove that $ \E[\hat{\mu}_1(t_i(k))] \ge {\mu_1\over 2} - 2\delta(T)$ with a probabilistic argument, where $\delta(T)$ converges to $0$ as $T$ goes to infinity. 
Thus, for large $\mu_1$ and small $\mu_2$ (so are $\mu_i$ for $i\geq2$), we have that $\E[\hat{\mu}_1(t_i(k))] - \mu_i= \Omega(\mu_1 - \mu_i)$ holds for any $i$ and $k \geq2$. This means that the compensation we need to pay for pulling arm $i$ once is about $\Theta(\mu_1 - \mu_i) = \Omega(\Delta_i)$. Thus, the total compensation $\Omega\left(\sum_{i=2}^N {\Delta_i\log T\over KL(D_i,D_1)}\right)$. $\Box$



\section{Compensation upper bound}

In this section, we consider three algorithms that can be applied to solve the KCMAB problem and present their analyses. Specifically, we consider the Upper Confidence Bound (UCB) Policy \citep{Auer2002Finite},  and propose a modified $\varepsilon$-Greedy Policy and a modified-Thompson Sampling Policy. 
Note that while the algorithms have been extensively analyzed for their regret performance, the compensation metric is significantly different from regret. Thus, the analyses are different and require new arguments.  

\subsection{The Upper Confidence Bound policy}
We start with the UCB policy shown in Algorithm \ref{Algorithm_UCB}. In the view of the long-term player, Algorithm \ref{Algorithm_UCB} is almost the same as the UCB policy in \cite{Auer2002Finite}, and its regret has been proved to be $O\left(\sum_{i=2}^N{\log T\over \Delta_i}\right)$. Thus,  we focus on the  compensation upper bound, which is shown in Theorem \ref{Theorem_UCB}. 

\begin{algorithm}[t]
    \centering
    \caption{The UCB algorithm for KCMAB.}\label{Algorithm_UCB}
    \begin{algorithmic}[1]
    \FOR {$t = 1,2,\cdots, N$}
    \STATE Choose arm $a(t) = t$.
    \ENDFOR
    \FOR {$t=N+1,\cdots$}
    \STATE For all arm $i$, compute $r_i(t) = \sqrt{2\log t\over N_i(t)}$ and $u_i(t) = \hat{\mu}_i(t) + r_i(t)$
    \STATE Choose arm $a(t) = \argmax_i u_i(t)$ (with compensation $\max_j \hat{\mu}_j(t) - \hat{\mu}_{a(t)}(t)$) 
    \ENDFOR
    \end{algorithmic}
\end{algorithm}

\begin{theorem}\label{Theorem_UCB}
	In Algorithm \ref{Algorithm_UCB}, we have that
	\begin{equation*}
		Com(T) \le \sum_{i=2}^N {16\log T\over \Delta_i} + {2N\pi^2\over 3}
	\end{equation*}
\end{theorem}

\textbf{Proof Sketch}: First of all,  it can be shown that the each sub-optimal arm is pulled for at most ${8\over \Delta_i^2}\log T$ times in Algorithm \ref{Algorithm_UCB} with high probability.  
Since in every time slot $t$ the long-term controller choose the arm $a(t) = \argmax_j \hat{\mu}_j(t) + r_j(t)$,  we must have $\hat{\mu}_{a(t)}(t) + r_{a(t)}(t) = \max_j (\hat{\mu}_j(t)+ r_j(t))\geq \max_j \hat{\mu}_j(t)$. This implies that the compensation is at most $r_{a(t)}(t)$. Moreover, if arm $a(t)$ has been pulled the maximum number of times, i.e., $N_{a(t)}(t) = \max_j N_j(t)$,   then $r_{a(t)}(t) = \min_j r_j(t)$ (by definition). Thus, $\hat{\mu}_{a(t)}(t) = \max_j \hat{\mu}_j(t)$, which means that the controller does not need to pay any compensation. 

Next, for any sub-optimal arm $i$, with high probability the compensation that the long-term controller pays for it can be upper bounded by:    
\begin{equation*}
Com_i(T) \leq \E\left[\sum_{\tau=1}^{N_i(T)} \sqrt{2\log T\over \tau}\right] \leq \E\left[\sqrt{8N_i(T)\log T}\right] \stackrel{(a)}{\le} \sqrt{8\E[N_i(T)]\log T} \le {8\log T \over \Delta_i}
\end{equation*}
Here the inequality (a) holds because $\sqrt{x}$ is concave. 
As for the optimal arm, when $N_1(t) \ge \sum_{i=2}^N{8\log T\over \Delta_i^2}$, with high probability $N_{1}(t) = \max_j N_j(t)$. Thus, the controller does not need to pay compensation in time slots with $a(t) = 1$ and $N_1(t) \ge \sum_{i=2}^N{8\log T\over \Delta_i^2}$. 
Using the same argument, the  compensation for arm $1$ is upper bounded by  $Com_1(T) \le \sum_{i=2}^N{8\log T\over \Delta_i}$ with high probability. 
Therefore, the overall compensation upper bound is given by $Com(T) \le \sum_{i=2}^N{16\log T\over \Delta_i}$ with high probability. $\Box$

\subsection{The modified $\varepsilon$-greedy policy}
The second algorithm we propose is a modified $\varepsilon$-greedy policy, whose details are presented in Algorithm \ref{Algorithm_Epoch}.  
The modified $\varepsilon$-greedy algorithm, though appears to be similar to the classic $\varepsilon$-greedy algorithm, has a critical difference. 
In particular, instead of randomly choosing an arm to explore, we use the round robin 
method to explore the arms. This guarantees that, given the number of total explorations, each arm will be explored a deterministic number of times. 
This facilitates the analysis for compensation upper bound. 


\begin{algorithm}[t]
    \centering
    \caption{The modified $\varepsilon$-greedy algorithm for KCMAB.}\label{Algorithm_Epoch}
    \begin{algorithmic}[1]
    \STATE  \textbf{Input: }$\epsilon$,
    \FOR {$t = 1,2,\cdots, N$}
    \STATE Choose arm $a(t) = t$.
    \ENDFOR
    \STATE $a_e \gets 1$
    \FOR {$t=N+1,\cdots$}
    \STATE With probability $\min\{1,{\epsilon\over t}\}$, choose arm $a(t) = a_e $ and set $a_e\gets (a_e\mod N) + 1$ (with compensation $\max_j \hat{\mu}_j(t) - \hat{\mu}_{a(t)}(t)$). 
    \STATE Else, choose the arm $a(t) = \argmax_i \hat{\mu}_i(t)$.
    \ENDFOR
    \end{algorithmic}
\end{algorithm}

In the regret analysis of the $\varepsilon$-greedy algorithm, the random exploration ensures that at  time slot $t$, the expectation of explorations on each arm is about ${\epsilon\over N} \log t$.  Thus, the probability that its empirical mean has a large error is small. In our algorithm, the number of explorations of each single arm is almost the same as classic $\varepsilon$-greedy algorithm in expectation (with only a small constant difference). Hence, adapting the analysis from $\varepsilon$-greedy algorithm gives the same regret upper bound, i.e. $O(\sum_{i=2}^N{\Delta_i\log T\over \Delta_2^2})$ when $\epsilon = {cN\over \Delta_2^2}$.  

Next,  we provide a compensation upper bound for our modified $\varepsilon$-greedy algorithm. 

\begin{theorem}\label{Theorem_Epoch}
    In Algorithm \ref{Algorithm_Epoch}, if we have $\epsilon = {cN\over \Delta_2^2}$, then
    \begin{equation}
    		Com(T) \le \sum_{i=2}^N {c\Delta_i\log T\over \Delta_2^2} + {N^2\over 2\Delta_2}\sqrt{c\log T}. \label{eq:epsilon-greedy-bdd}
    \end{equation}
\end{theorem}

\textbf{Proof Sketch:}  Firstly, our modified $\varepsilon$-greedy algorithm chooses the arm with the  largest empirical mean in non-exploration steps. Thus, we only need to consider  the exploration steps, i.e.,  steps during which we choose to explore arms according to round-robin. Let $t_i^{\varepsilon}(k)$ be the time slot that we explore arm $i$ for the $k$-th time. Then the compensation the controller has to pay in this time slot is $\E[\max_j \hat{\mu}_j(t_i^{\varepsilon}(k)) - \hat{\mu}_i(t_i^{\varepsilon}(k))]$. 

Since the rewards are independent of whether we choose to explore, one sees that $\E[\hat{\mu}_i(t_i^{\varepsilon}(k))] = \mu_{i}$. Thus, we can decompose $\E[\max_j \hat{\mu}_j(t_i^{\varepsilon}(k)) - \hat{\mu}_i(t_i^{\varepsilon}(k))]$ as follows: 
\begin{eqnarray}
	\E[\max_j \hat{\mu}_j(t_i^{\varepsilon}(k)) - \hat{\mu}_i(t_i^{\varepsilon}(k))] & = &\E[\max_j (\hat{\mu}_j(t_i^{\varepsilon}(k)) - \mu_{i})] \nonumber\\
	&\le& \E[\max_j (\hat{\mu}_j(t_i^{\varepsilon}(k)) - \mu_j)] + \E[\max_j (\mu_j - \mu_{i})]. \label{eq:epsilon-greedy-step0}
\end{eqnarray}
The second term in \eqref{eq:epsilon-greedy-step0} is bounded by $\Delta_{i} = \mu_1 - \mu_{i}$. Summing over all these steps and all arms, we obtain the first term $\sum_{i=2}^N {c\Delta_i\log T\over \Delta_2^2}$  in our bound \eqref{eq:epsilon-greedy-bdd}. 

We turn to the first term in \eqref{eq:epsilon-greedy-step0}, i.e.,  $\E[\max_j (\hat{\mu}_j(t_i^{\varepsilon}(k)) - \mu_j)]$. We see that it is  upper bounded by
\begin{equation*}
\E[\max_j (\hat{\mu}_j(t_i^{\varepsilon}(k)) - \mu_j)] \le \E[\max_j (\hat{\mu}_j(t_i^{\varepsilon}(k)) - \mu_j)^+] \le \sum_j \E[(\hat{\mu}_j(t_i^{\varepsilon}(k)) - \mu_j) ^+]
\end{equation*}
where $(*)^+ = \max\{*,0\}$.
When arm $i$ has been explored $k$ times (line $7$ in Algorithm \ref{Algorithm_Epoch}), we know that all other arms have at least $k$ observations (in the first $N$ time slots, there is one observation for each arm). Hence, $\E[(\hat{\mu}_j(t_i^{\varepsilon}(k)) - \mu_j) ^+] = {1\over 2}\E[|\hat{\mu}_j(t_i^{\varepsilon}(k)) - \mu_j|]\le {1\over 4\sqrt{k}}$ (the equality is due to the fact that $\E[|x|]=2\E[x^+]$ if $\E[x]=0$). 

Suppose arm $i$ is been explored in time set $T_i = \{t_i^{\varepsilon}(1),\cdots\}$. Then, \begin{equation*}
\sum_{k \le |T_i|} \E[\max_j (\hat{\mu}_j(t_i^{\varepsilon}(k)) - \mu_j)^+] \le \sum_{k \le |T_i|} {N\over 4\sqrt{k}} \le {N\sqrt{|T_i|}\over 2}
\end{equation*}
Since $\E[|T_i|] = {c\over \Delta_2^2}\log T$, we can bound the first term in \eqref{eq:epsilon-greedy-step0} as ${N^2\sqrt{c\log T}\over 2\Delta_2}$. Summing this with $\sum_{i=2}^N {c\Delta_i\log T\over \Delta_2^2}$ above for the second term, we obtain  the compensation upper bound in (\ref{eq:epsilon-greedy-bdd}). $\Box$ 

\subsection{The Modified Thompson Sampling policy}
The third algorithm we propose is a Thompson Sampling (TS) based policy. 
Due to the complexity of the analysis for the traditional TS algorithm,  we propose  a modified TS policy  and derive its compensation bound. 
Our modification is motivated by the idea of the LUCB algorithm \citep{kalyanakrishnan2012pac}. 
Specifically,  we divide time into rounds containing two time steps each, and pull not only the arm with largest sample value, but also the arm with largest empirical mean in each round.  
%
The modified TS policy is presented in Algorithm \ref{Algorithm_TS}, and we have the following theorem about its regret and compensation. 

\begin{algorithm}[t]
    \centering
    \caption{The Modified Thompson Sampling Algorithm for KCMAB.}\label{Algorithm_TS}
    \begin{algorithmic}[1]
    \STATE  \textbf{Init: }$\alpha_i = 1, \beta_i = 1$ for each arm $i$.
    \FOR {$t = 1,2,\cdots, N$}
    \STATE Choose arm $a(t) = t$ and receive the observation $X(t)$.
    \STATE {\sf Update}$(\alpha_{a(t)},\beta_{a(t)},X(t))$
    \ENDFOR
    \FOR {$t=N+1,N+3,\cdots$}
    \STATE For all $i$ sample values $\theta_i(t)$ from Beta distribution $\B(\alpha_i,\beta_i)$;
    \STATE Play action $a_1(t) = \argmax_i \hat{\mu}_i(t)$, get the observation $X(t)$. {\sf Update}$(\alpha_{a_1(t)},\beta_{a_1(t)},X(t))$
    \STATE Play action $a_2(t+1) = \argmax_i \theta_i(t)$ (with compensation $\max_j \hat{\mu}_j(t+1) - \hat{\mu}_{a_2(t+1)}(t+1)$), receive the observation $X(t+1)$. {\sf Update}$(\alpha_{a_2(t+1)},\beta_{a_2(t+1)},X(t+1))$
    \ENDFOR
    \end{algorithmic}
\end{algorithm}

\begin{algorithm}[t]
    \centering
    \caption{Procedure {\sf Update}}\label{Update}
    \begin{algorithmic}[1]
    \STATE \textbf{Input:} $\alpha_i,\beta_i,X(t)$
    \STATE \textbf{Output:} updated $\alpha_i,\beta_i$
    \STATE $Y(t) \gets 1$ with probability $X(t)$, $0$ with probability $1-X(t)$
    \STATE $\alpha_i \gets \alpha_i+ Y(t)$; $\beta_i \gets \beta_i+1 - Y(t)$
    \end{algorithmic}
\end{algorithm}

\begin{theorem}\label{Theorem_TS}

In Algorithm \ref{Algorithm_TS}, we have
\begin{equation*}
	Reg(T) \le \sum_i {2\Delta_i\over (\Delta_i - \varepsilon)^2} \log T + O\left({N\over \varepsilon^4}\right) + F_1(\bm{\mu})
\end{equation*}
for some small $\varepsilon < \Delta_2$ and $F_1(\bm{\mu})$ does not depend on $(T,\varepsilon)$. As for compensation, we have:
\begin{equation*}
	Com(T) \le \sum_i {8 \over \Delta_i - \varepsilon} \log T + N\log T + O\left({N\over \varepsilon^4}\right) + F_2(\bm{\mu})
\end{equation*} where $F_2(\bm{\mu})$ does not depend on $(T,\varepsilon)$ as well.

\end{theorem}

\textbf{Proof Sketch:}  In round $(t, t+1)$, we assume that we first run the arm with largest empirical mean on time slot $t$ and call $t$ an empirical step. Then we run the arm with largest sample on time slot $t+1$ and call $t+1$ a sample step. 

We can bound the number of sample steps during which we pull a sub-optimal arm, using existing results in \cite{agrawal2013further}, since all  sample steps form an approximation of the classic TS algorithm. Moreover, \cite{Kaufmann2012Thompson} shows that in sample steps, the optimal arm is pulled for many times (at least $t^b$ at time $t$ with a constant $b \in (0,1)$). 
Thus, after several steps, the empirical mean of the optimal arm will be accurate enough. Then, if we choose to pull sub-optimal arm $i$ during empirical steps, arm $i$ must have an inaccurate empirical mean. Since the pulling will update its empirical mean, it is harder and harder for the arm's empirical mean to remain  inaccurate. As a result, it cannot be pulled for a lot of times during the empirical steps as well. 


Next, we discuss how to bound its compensation. It can be shown that with high probability, we always have $|\theta_i(t) - \hat{\mu}_i(t)| \le r_i(t)$, where $r_i(t) = \sqrt{2\log t \over N_i(t)}$ is defined in Algorithm \ref{Algorithm_UCB}.
Thus, we can focus on the case that $|\theta_i(t) - \hat{\mu}_i(t)| \le r_i(t)$ happens for any $i$ and $t$. 
Note that we do not need to pay compensation in empirical steps. In sample steps, suppose we pull arm $i$ and the largest empirical mean is in arm $j\neq i$ at the beginning of this round. Then, we need to pay $\max_k \hat{\mu}_k(t+1) - \hat{\mu}_{i}(t+1)$, which is upper bounded by $\hat{\mu}_j(t) - \hat{\mu}_i(t) + (\hat{\mu}_j(t+1)-\hat{\mu}_j(t))^+ \le \hat{\mu}_j(t) - \hat{\mu}_i(t) + {1\over N_j(t)}$ (here $\hat{\mu}_{i}(t+1)=\hat{\mu}_{i}(t)$). 
As $\theta_i(t) \ge \theta_j(t)$, we must have $\hat{\mu}_i(t) + r_i(t) \ge \theta_i(t) \ge \theta_j(t) \ge \hat{\mu}_j(t) - r_j(t)$, which implies $\hat{\mu}_j(t) - \hat{\mu}_{i}(t) \le r_i(t) + r_j(t)$. 
Thus, what we need to pay is at most $r_i(t) + r_j(t) + {1\over N_j(t)}$ if $i\ne j$, in which case we can safely assume that we pay $r_j(t) + {1\over N_j(t)} $ during empirical steps, and $r_i(t)$ during sample steps. 

For an sub-optimal arm $i$, we have $Com_i(T) \le \sum_i {4 \over \Delta_i - \varepsilon} \log T + O({1\over \varepsilon^4}) + F_1(\bm{\mu}) + \log T$ (summing over $r_i(t)$ gives the same result as in the UCB case, and   summing over ${1\over N_i(t)}$ is upper bounded by $\log T$).  
As for arm $1$, when $a_1(t) = a_2(t+1) = 1$, we do not need to pay $r_1(t)$ twice. In fact, we only need to pay at most ${1\over N_1(t)}$. Then, the number of time steps that $a_1(t) = a_2(t+1) = 1$ does not happen is upper bounded by $\sum_{i=2}^N \left({2\over (\Delta_i-\varepsilon)^2} \log T\right) + O\left({N\over \varepsilon^4}\right) + F_1(\bm{\mu})$, which is given by regret analysis. Thus, the compensation we need to pay on arm $1$ is upper bounded by $\sum_i {4 \over \Delta_i - \varepsilon} \log T + O({1\over \varepsilon^4}) + F_1(\bm{\mu}) + \log T$. Combining the above, we have the compensation bound $Com(T) \le \sum_i {8 \over \Delta_i - \varepsilon} \log T + N\log T+ O({1\over \varepsilon^4}) + F_2(\bm{\mu}) $. $\Box$

\section{Experiments}

In this section, we present experimental results for   the three algorithms, i.e., the UCB policy, the modified $\varepsilon$-greedy policy and the modified TS policy. We also compare our modified TS policy with origin TS policy to evaluate their difference. 
In our experiments, there are a total of nine arms with expected reward vector $\bm{\mu} = [0.9,0.8,0.7,0.6,0.5,0.4,0.3,0.2,0.1]$. We run the game for $T = 10000$ time steps. The experiment runs for $1000$ times and we take the average over these results. The ``-R'' represents the regret of that policy, and ``-C'' represents the compensation.

\begin{figure}
\begin{minipage}[t]{0.32\linewidth}
\centering
\includegraphics[width=1.7in]{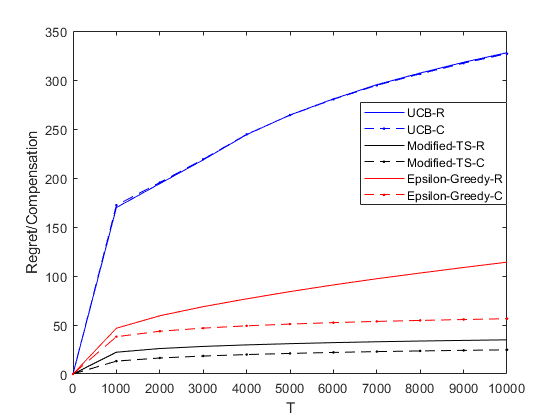}
\caption{Regret and Compensation of Three policies.}
\label{Figure_1}
\end{minipage}
\begin{minipage}[t]{0.32\linewidth}
\centering
\includegraphics[width=1.7in]{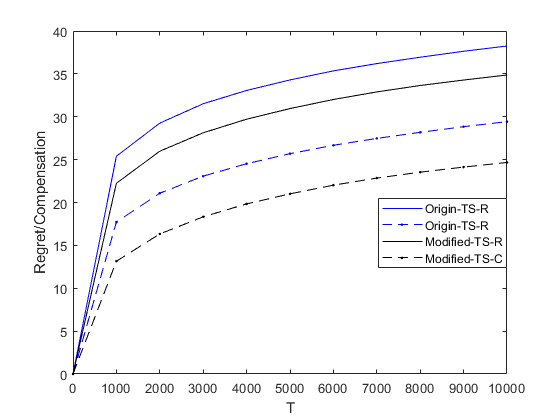}
\caption{Regret and Compensation of TS and modified-TS.}
\label{Figure_2}
\end{minipage}
\begin{minipage}[t]{0.32\linewidth}
\centering
\includegraphics[width=1.7in]{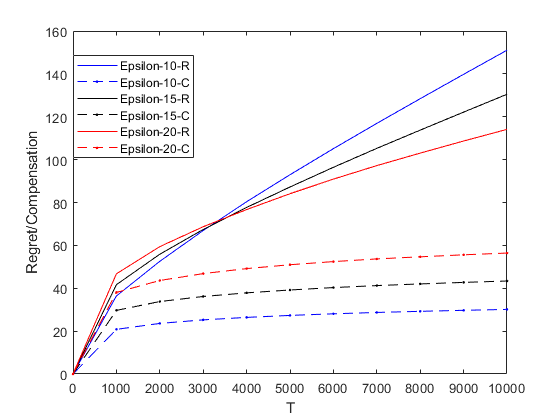}
\caption{Regret and Compensation of modified $\varepsilon$-greedy.}
\label{Figure_3}
\end{minipage}
\end{figure}

The comparison of the three policies in this paper is shown in Figure \ref{Figure_1}. We can see that modified-TS performs best in both regret and compensation, compared to other algorithms. 
As for the modified $\varepsilon$-greedy policy, when the parameter $\epsilon$ is chosen properly, it can also achieve a good performance. In our experiment, we choose $\epsilon = 20$. 

In Figure \ref{Figure_2}, we see that modified-TS performs better than TS in both compensation and regret, which means that our modification is effective. 
Figure \ref{Figure_3} shows the different performance of the modified $\varepsilon$-greedy policies with different $\epsilon$ values. Here we choose $\epsilon$ to be $10$,$15$ and $20$. From the experiments, we see the trade-off between regret and compensation: low compensation leads to high regret, and high compensation leads to low regret. 


\section{Conclusion}

We propose and study the KCMAB   problem where a controller offers compensation to incentivize players for arm exploration. We first provide the analysis of a compensation lower bound achieved by regret-minimizing algorithms. 
Then, we consider three algorithms, namely, UCB,  modified $\varepsilon$-greedy and modified TS. We show that all three algorithms achieve good regret bounds, while keeping  order-optimal  compensation. 
We also conduct experiments and the results validate our theoretical findings. 
%

\clearpage

\bibliography{general}
\bibliographystyle{abbrvnat}

\onecolumn

\appendix

\section*{Supplementary Material}


	


The proofs of all lemmas are shown in the end of the sections they belong to.

\section{Proof of Theorem \ref{Theorem_LB}}

\begin{fact}\label{Fact_Lai}(Theorem 2 in \cite{Lai1985Asymptotically})
If an algorithm guarantees $o(T^{\alpha})$ regret upper bound for any $\alpha >0$, then for any $\varepsilon > 0$, we have:
\begin{equation*}
  \lim_{T\to \infty} \Pr[N_i(t) \ge (1-\varepsilon){\log T\over KL(D_i,D_1)}] = 1
\end{equation*}
\end{fact}

Fact \ref{Fact_Lai} means that we will pull arm $i$ for at least $(1-\varepsilon){\log T\over KL(D_i,D_1)}$ times almost surely, i.e. there exists a function $\delta(T) \to 0$ as $T \to \infty$ such that $\Pr[N_i(T) \ge (1-\varepsilon){\log T\over KL(D_i,D_1)}] \ge 1 - \delta(T)$.

Consider the time step when we pull arm $i$ for the $k$-th time ($2 \le k \le (1-\varepsilon){\log T\over KL(D_i,D_1)}$), and we use $t_i(k)$ to denote the random variable of this time slot. Since we may not pull arm $i$ for $k$ times until time $T$, we suppose that the game lasts for infinite number of times, and $t_i(k)$ can be larger than $T$. By the definition of $t_i(k)$, we must have $\E[\hat{\mu}_i(t_i(k))] = \mu_i$. 

Then the compensation we need to pay for pulling arm $i$ for the $k$-th time can be bounded as following:

\begin{eqnarray}
	\nonumber\Pr[t_i(k) \le T]\E[c(t_i(k)) | t_i(k) \le T] &=& \Pr[t_i(k) \le T]\E[\max_j \hat{\mu}_j(t_i(k)) - \hat{\mu}_i(t_i(k)) | t_i(k) \le T]\\
	\label{eq_m1}&\ge& \E[\max_j \hat{\mu}_j(t_i(k)) - \hat{\mu}_i(t_i(k))] - \Pr[t_i(k) > T]\\
	\label{eq_m2}&\ge& \E[\max_j \hat{\mu}_j(t_i(k))] - \E[\hat{\mu}_i(t_i(k))] - \delta(T)\\
	\nonumber&\ge& \E[\hat{\mu}_1(t_i(k))] - \E[\hat{\mu}_i(t_i(k))] - \delta(T)\\
	\label{eq_m12}&=&  \E[\hat{\mu}_1(t_i(k))] - \mu_i - \delta(T)
\end{eqnarray}

Eq. \eqref{eq_m1} is because that $\max_j \hat{\mu}_j(t_i(k)) - \hat{\mu}_i(t_i(k)) \le 1$, and Eq. \eqref{eq_m2} is because that $\Pr[t_i(k) > T] \le \delta(T)$, which is given by Fact \ref{Fact_Lai}.




In Theorem \ref{Theorem_LB}, we suppose that $D_1$ is a Bernoulli distribution, i.e. $\Pr[X_1(t) = 1] = \mu_1$ and $\Pr[X_1(t) = 0] = 1-\mu_1$. Then we can use a 0-1 string to represent the history of arm $1$. In this case $\hat{\mu}_1(t_i(k)) = {\#(s)\over |s|}$, where $\#(s)$ is the number of $1$s in string $s$. To simplify the notations, we use $z-s$ to denote the string that removes prefix $s$ from $z$, and $s+z$ to denote the string given by adding prefix $s$ to $z$.

We can thus rewrite $\E[\hat{\mu}_1(t_i(k))]$ as:
\begin{equation*}
	\E[\hat{\mu}_1(t_i(k))] = \sum_{s} p(s){\#(s)\over |s|},
\end{equation*}
where $p(s)$ is the probability that at time slot $t_i(k)$, the history of arm $1$ forms string $s$. Note that this expectation is hard to evaluate, since all the arms are coupled due to the algorithms, which makes $\hat{\mu}_1(t_i(k))$ dependent on $t_i(k)$.

We define two events as following: $\mathcal{A}_L(s)$ is the event that the first $|s|$ feedbacks of arm $1$ form string $s$, and $\mathcal{B}_L(s)$ is the event that at time slot $t_i(k)$, the feedbacks of arm $1$ forms string $s$. Then $\Pr[\mathcal{A}_L(s)] = P(s,\mu_1)$, where $P(s,\mu_1) = \mu_1^{\#(s)}(1-\mu_1)^{|s| - \#(s)} $, and $\Pr[\mathcal{B}_L(s)] = p(s)$.
%

In our model, we suppose that we pull each arm once in the first $N$ time slots, then $\sum_{|s| = 0} p(s) = 0$. By these two equations, we have $\sum_{|s| > 0} p(s) = 1$, thus $p$ is a probability distribution on all 0-1 strings.
Since $p$ is a probability distribution, when the first $|s|$ feedbacks of arm $1$ form string $s$, there must be a string $z$ such that $\mathcal{B}_L(z)$ happens and either $z$ is prefix of $s$ or $s$ is prefix of $z$. Moreover, if $z$ is prefix of $s$, we still need the next $|s| - |z|$ feedbacks from arm 1 form the string $s-z$. This means the following equation holds:
\begin{equation}
\mathcal{A}_L(s) = \cup_{z: pre(z,|s|) = s} \mathcal{B}_L(z) \cup_{y\in sub(s)} (\mathcal{B}_L(y) \cap \{\text{The next }|s|-|y|\text{ feedbacks form string }s-y\}),
\end{equation}
where $pre(z,n)$ is the prefix of $z$ with length $n$, and $sub(s) = \{pre(s,j)|1\le j \le |s|-1\}$ is the set that contains all prefixes of $s$ but does not contain $s$ itself.

For any $s\ne z$, we mush have $\mathcal{B}_L(s) \cap \mathcal{B}_L(z) = \emptyset$, thus

\begin{eqnarray}
	\nonumber\Pr[\mathcal{A}_L(s)] &=& \sum_{z: pre(z,|s|) = s}\Pr[\mathcal{B}_L(z)] + \sum_{y\in sub(s)} \Pr[\mathcal{B}_L(y) \cap \{\text{The next }|s|-|y|\text{ feedbacks forms string }s-y\}]\\
	\label{eq_m3}&=&\sum_{z: pre(z,|s|) = s} p(z) + \sum_{y\in sub(s)} p(y)P(s-y, \mu_1)
\end{eqnarray}

Eq. \eqref{eq_m3} is because that the next $|s|-|y|$ feedbacks are independent with event $\mathcal{B}_L(y)$. From this equation, we have $P(s,\mu_1) = \sum_{z: pre(z,|s|) = s} p(z) + \sum_{y\in sub(s)} p(y)P(s-y, \mu_1) $.

To simplify the analysis, we construct $p^{T}$ from $p$ by adding the probability of $p(s)$ with $s > T$ to $p(pre(s,T))$:

\[
p^{T}(s)=
\begin{cases}p(s) \quad \ \ &
 1\le|s|<T\\
p(s) + \sum_{z: pre(z,T) = s} p(z) \quad \ \ &
|s|=T\\
0 \quad \ \ &|s|>T
\end{cases}
\]

This does not influence the Eq. \eqref{eq_m3} for all $s$ with $|s|\leq T$ (the probability mass for strings longer than $T$ is included in its $T$-sized prefix). Thus $P(s,\mu_1) = \sum_{z: pre(z,|s|) = s} p^{T}(z) + \sum_{y\in sub(s)} p^{T}(y)P(s-y, \mu_1) $ still holds.

Now we can use $p^{T}$ to bound $\E[\hat{\mu}_1(t_i(k))]$ as:

\begin{equation}\label{eq_m10}
\E[\hat{\mu}_1(t_i(k))] = \sum_s p(s){\#(s)\over |s|} \ge \sum_s p^{T}(s){\#(s)\over |s|} - \delta(T)
\end{equation}
%
Here  
 $\sum_{|s| > T} p(s) \le \delta(T)$ holds by Fact \ref{Fact_Lai} ($\cup_{|s| > T} \mathcal{B}_L(s)$ implies $N_i(T)\leq k\leq  (1-\varepsilon){\log T\over KL(D_i,D_1)}$).

From $p^{T}$, we can then build another $q^{T}$ as follows. 

\[
q^{T}(s)=
\begin{cases}{p^{T}(s)\over P(s,\mu_1)  - \sum_{y\in sub(s)}p^{T}(y) P(s-y,\mu_1)} \quad \ \ &
P(s,\mu_1)  - \sum_{y\in sub(s)}p^{T}(y) P(s-y,\mu_1) > 0\\
1 \quad \ \ &
P(s,\mu_1)  - \sum_{y\in sub(s)}p^{T}(y) P(s-y,\mu_1) = 0
\end{cases}
\]

One can check that $q^{T}(s) = 1$ for any $|s| = T$ and  $0\le q^{T}(s) \le 1$ for any $1 \le |s| < T$.

\begin{lemma}\label{Lemma_LB_8}
We can get the value of $p^{T}$ from $q^{T}$ by the following equation, 
\begin{equation}\label{eq_m31}
p^{T}(s) = P(s,\mu_1) \prod_{j=1}^{|s|-1} (1-q^{T}(pre(s,j))) q^{T}(s)
\end{equation}
\end{lemma}

Lemma \ref{Lemma_LB_8} means that every possible $p^{T}$ have a unique $q^{T}$ match and vice versa. Then we can consider a set of $q^{T}(s)$ such that the corresponding $p^{T}(s)$ minimizes $ \sum_{ 1\le |s| \le T} p^{T}(s) {\#(s) \over |s|}$. We write $emp(q^{T},\mu_1, T)$ to denote the corresponding value $ \sum_{ 1\le |s| \le T} p^{T}(s) {\#(s) \over |s|}$. Then for given $q^{T}$, we can use Algorithm \ref{Algorithm_cal_emp} to compute $emp(q^{T},\mu_1, T) $.


\begin{algorithm}[t]
    \centering
    \caption{Calculate $emp(q^{T},\mu_1, T)$}\label{Algorithm_cal_emp}
    \begin{algorithmic}[1]
    \STATE  \textbf{Input: } $q^{T}$, $\mu_1$, $T$.
    \FOR {$s$ with $|s| = T$}
    \STATE $g(s) \gets {\#(s) \over |s|}$.
    \ENDFOR
    \FOR {$t = T-1,\cdots,1$}
    \FOR {$s$ with $|s| = t$}
    \STATE $g(s) \gets q^{T}(s){\#(s) \over |s|} + (1-q^{T}(s))(\mu g(S+\text{"1"}) + (1-\mu)g(S+\text{"0"}))$
    \ENDFOR
    \ENDFOR
    \STATE $emp(q^{T},\mu_1, T) = \mu g(\text{"1"}) + (1-\mu) g(\text{"0"})$
    \STATE \textbf{Output: } $emp(q^{T},\mu_1, T)$.
    \end{algorithmic}
\end{algorithm}

\begin{proposition}\label{Propo_corr}
  Algorithm \ref{Algorithm_cal_emp} returns the value $emp(q^{T},\mu_1, T)$ correctly.
\end{proposition}

To find $q^{T}$ that minimizes $emp(q^{T},\mu_1, T)$, we introduce a Dynamic Programming policy as Algorithm \ref{Algorithm_DP_emp}. It starts by setting $f(s) = {\#(s)\over |s|}$ for all strings with $|s| = T$. After that, if all strings $s$ with $|s| = k$ have their $f(s)$, we start to consider $s$ with $|s| = k-1$. For any $|s| = k-1$, the DP policy will check whether it is good or not to stop at $s$, i.e. only if ${\#(s)\over |s|}$ is smaller than $\mu_1 f(s+\text{"1"}) + (1-\mu_1)f(s+\text{"0"})$, we choose to set $q^{T}(s) = 1$ and $f(s) = {\#(s)\over |s|}$, otherwise we set $q^{T}(s) = 0$ and $f(s) = \mu_1 f(s+\text{"1"}) + (1-\mu_1)f(s+\text{"0"})$.


Intuitively, the DP policy is the best one can do, which is shown in the following lemma.

\begin{algorithm}[t]
    \centering
    \caption{$DP$ stopping policy}\label{Algorithm_DP_emp}
    \begin{algorithmic}[1]
    \STATE  \textbf{Input: } $\mu_1$, $T$.
    \FOR {$a=0,1,\cdots,T$}
    \STATE  $f(a,T-a) \gets {a\over T}$.
    \ENDFOR
    \FOR {$t = T-1,\cdots,1$}
    \FOR {$a=0,1,\cdots,t$}
    \STATE $f(a,t-a) \gets \min \{{a\over t}, \mu f(a+1,t-a) + (1-\mu) f(a,t-a+1)\}$
    \ENDFOR
    \ENDFOR
    \STATE $f(0,0) = \mu f(1,0) + (1-\mu) f(0,1)$
    \STATE \textbf{Output: } $DP(\mu_1,T) = f(0,0)$ 
    \end{algorithmic}
\end{algorithm}

\begin{lemma}\label{Lemma_DP_opt}
  For any given $q^{T}$, $emp(q^{T}, \mu_1, T) \ge DP(\mu_1,T)$, where $DP(\mu_1,T)$ is the output value of Algorithm \ref{Algorithm_DP_emp}.
\end{lemma}

Now from Eq. \eqref{eq_m10} and Lemma \ref{Lemma_DP_opt}, we have:

\begin{equation}\label{eq_m11}
\E[\hat{\mu}_1(t_i(k))]  \ge  DP(\mu_1,T) - \delta(T)
\end{equation}

Then we need a lower bound on $DP(\mu_1,T)$, which is given in the following lemma.

\begin{lemma}\label{Lemma_min_emp}
  For any $\mu_1 \ge 0.9$ and $T\ge 1$, we have $DP(\mu_1,T) \ge {\mu_1\over 2}$
\end{lemma}

From Lemma \ref{Lemma_min_emp}, Eq. \eqref{eq_m11} and Eq. \eqref{eq_m12}, we know that when $\mu_1 \ge 0.9$, $0.2 \ge \mu_2\ge \mu_3 \ge \cdots \ge \mu_N$, $\E[c(t_i(k))] \ge DP(\mu_1,T) -\mu_i - 2\delta(T) \ge {\Delta_i\over 4} - 2\delta(T)$ for any $i$ and $2\le k \le (1-\varepsilon){\log T\over KL(D_i,D_1)}$. Thus the compensation we need to pay is $\Omega(\sum_{i=2}^N (\Delta_i - 2\delta(T)){\log T\over KL(D_i,D_1)})$. When $T \to \infty$, we have $\delta(T) \to 0$. Therefore, the compensation is lower bounded by $\Omega(\sum_{i=2}^N {\Delta_i \log T\over KL(D_i,D_1)})$.

\subsection{Proof of Lemma \ref{Lemma_LB_8}}

We prove this lemma by induction.

For the strings $s$ with $|s| = 1$, since there is no string in $sub(s)$, we have $q^{T}(s) = {p^T(s)\over P(s,\mu_1)}$ by definition of $q^{T}$. Thus Eq. \eqref{eq_m31} holds.

If for all strings $s$ with $|s| \le k$, Eq. \eqref{eq_m31} holds. Then consider a string $s$ with $|s| = k + 1$, we choose $z$ as the longest string in $sub(s)$ such that $q^{T}(z) > 0$.

If such $z$ does not exist. Then by definition of $q^{T}(z)$, we know $p^{T}(z) = 0$ for all $z \in sub(s)$. Thus 
we have $P(s,\mu_1)\prod_{j=1}^{|s|-1} (1-q^{T}(pre(s,j))) q^{T}(s) = P(s,\mu_1)q^{T}(s) = p^{T}(s)$ holds.

When such $z$ exists, let $x = s - z$, and then we have

\begin{eqnarray}
\nonumber&&P(s,\mu_1)\prod_{j=1}^{|s|-1} (1-q^{T}(pre(s,j))) q^{T}(s) \\
\nonumber=&& \left(P(z,\mu_1)\prod_{j=1}^{|z|-1} (1-q^{T}(pre(z,j)))\right)(1-q^{T}(z))q^{T}(s)P(x,\mu_1)\\
\label{eq_m41}=&& {p^{T}(z)\over q^{T}(z)}(1-q^{T}(z))q^{T}(s)P(x,\mu_1)
\end{eqnarray}

Eq. \eqref{eq_m41} is because that by induction, Eq. \eqref{eq_m31} holds for any $|z| \le k$.

If $P(z,\mu_1)  - \sum_{y\in sub(z)}p^{T}(y) P(z-y,\mu_1) = 0$, then we must have $p^T(z) = 0$, thus $P(s,\mu_1)  - \sum_{y\in sub(s)}p^{T}(y) P(s-y,\mu_1) \le P(x,\mu_1)\left(P(z,\mu_1)  - \sum_{y\in sub(z)}p^{T}(y) P(z-y,\mu_1)\right) = 0$, which means $p^T(s) = 0$. On the other hand, $P(z,\mu_1)  - \sum_{y\in sub(z)}p^{T}(y) P(z-y,\mu_1) = 0$ means $q^{T}(z) = 1$, thus Eq. \eqref{eq_m31} holds.

If $P(z,\mu_1)  - \sum_{y\in sub(z)}p^{T}(y) P(z-y,\mu_1) > 0$, then 
\begin{eqnarray}
\nonumber&& {p^{T}(z)\over q^{T}(z)}(1-q^{T}(z))q^{T}(s)P(x,\mu_1) \\
\nonumber =&& p^{T}(z){1-q^{T}(z)\over q^{T}(z)}q^{T}(s)P(x,\mu_1)\\
\nonumber= && p^{T}(z) \left({P(z,\mu_1)  - \sum_{y\in sub(z)}p^{T}(y) P(z-y,\mu_1) \over p^{T}(z)} - 1\right)q^{T}(s)P(x,\mu_1)\\
\nonumber=&&\left(P(z,\mu_1)  - \sum_{y\in sub(z)}p^{T}(y) P(z-y,\mu_1) - p^{T}(z)\right)q^{T}(s)P(x,\mu_1)\\
\label{eq_m42}=&&\left(P(s,\mu_1)  - \sum_{y\in sub(s)}p^{T}(y) P(s-y,\mu_1)\right)q^{T}(s)\\
\nonumber=&&p^{T}(s)
\end{eqnarray}

Eq. \eqref{eq_m42} is because that for all $s' \in sub(s)$ and $|s'|>|z|$, $q^{T}(s') = 0$ implies $p^{T}(s') = 0$.

By induction, we finish the proof of Lemma \ref{Lemma_LB_8}.

\subsection{Proof of Lemma \ref{Lemma_DP_opt}}
  
We use induction to prove that $emp(q^{T},\mu_1, T) \ge DP(\mu_1,T)$.
  
  For all $s$ with size $|s| = T$, we can see that $f(\#(s), |s| - \#(s)) = {\#(s) \over |s|} \le g(s)$, where $f(\#(s), |s| - \#(s))$ is the value in Algorithm \ref{Algorithm_DP_emp} with input $\mu_1,T$.
  
  If for all $s$ with size $|s| = k$, we have $f(\#(s), |s| - \#(s)) \le g(s)$, then consider any $s'$ with size $|s'| = k-1$.
  
  \begin{eqnarray*}
    f(\#(s'), |s'| - \#(s')) &=& \min\{{\mathbf{1}(s') \over |s'|}, \mu f(\#(s)+1, |s| - \#(s)) + (1-\mu)f(\#(s), |s| - \#(s)+1)\} \\
    &\le& \min\{{\mathbf{1}(s') \over |s'|}, \mu g(s'+\text{"1"}) + (1-\mu)g(s'+\text{"0"})\} \\
    &\le& g(s')
  \end{eqnarray*}
  
  Thus by induction, we have $f(0,1) \le g(\text{"0"})$ and $f(1,0) \le g(\text{"1"})$, which means $emp(q^{T},\mu_1, T) \ge DP(\mu_1,T)$.

\subsection{Proof of Lemma \ref{Lemma_min_emp}}

  We calculate this by summing over the difference between $DP(\mu_1,T)$ and $DP(\mu_1,T+1)$. Let $f^{\mu_1,T}(a,b) - f^{\mu_1,T+1}(a,b) = \delta^{\mu_1,T}_{a,b}$, where $f^{\mu,T}(a,b)$ is the value of $f(a,b)$ when inputting $\mu,T$ into Algorithm \ref{Algorithm_DP_emp}.
  
  First consider the case that $a+b = T$, $f^{\mu_1,T}(a,b) = {a\over T}$, while $f^{\mu_1,T+1}(a,b) = \min\{{a\over T}, \mu_1{a+1\over T+1} + (1-\mu_1){a\over T+1}\}$.
  
  Since $\mu_1{a+1\over T+1} + (1-\mu_1){a\over T+1} = {a+\mu_1\over T+1} = {a\over T} + {\mu_1 - a/T \over T+1} = {a\over T} + {\mu_1 T -a \over T(T+1)}$. When $a > \mu_1 T$, we have $\delta^{\mu_1,T}_{a,b} = {a-\mu_1 T \over T(T+1)}$; otherwise $\delta^{\mu_1,T}_{a,b} = 0$. Thus $\delta^{\mu_1,T}_{a,b} = {1\over T(T+1)} (a-\mu_1 T)^+$.
  
  Then we consider the case that $a + b = t < T$. By definition, $f^{\mu_1,T}(a,b) = \min\{{a\over t}, \mu_1 f^{\mu_1,T}(a+1,b) + (1-\mu_1)f^{\mu_1,T}(a,b+1)\}$. Thus 
  
  \begin{eqnarray}
    \nonumber\delta^{\mu_1,T}_{a,b} &=& f^{\mu_1,T}(a,b) - f^{\mu_1,T+1}(a,b) \\
    \nonumber&=& \min\{{a\over t}, \mu_1 f^{\mu_1,T}(a+1,b) + (1-\mu_1)f^{\mu_1,T}(a,b+1)\} \\
    \nonumber&&-\min\{{a\over t}, \mu_1 f^{\mu_1,T+1}(a+1,b) + (1-\mu_1)f^{\mu_1,T+1}(a,b+1)\} \\
    \nonumber&\le& (\mu_1 f^{\mu_1,T}(a+1,b) + (1-\mu_1)f^{\mu_1,T}(a,b+1))\\
    \label{eq_m21}&& - (\mu_1 f^{\mu_1,T+1}(a+1,b) + (1-\mu_1)f^{\mu_1,T+1}(a,b+1)) \\
    \nonumber&=& \mu_1 \delta^{\mu_1,T}_{a+1,b} + (1-\mu_1)\delta^{\mu_1,T}_{a,b+1}
  \end{eqnarray}
  
  Eq. \eqref{eq_m21} is because of the fact that $f^{\mu_1,T}(a,b) \ge f^{\mu_1,T+1}(a,b)$ for any given $(a,b)$.
  
  This implies :
  
  \begin{eqnarray*}
    \delta^{\mu_1,T}_{0,0} &\le& \mu_1 \delta^{\mu_1,T}_{1,0} + (1-\mu_1) \delta^{\mu_1,T}_{0,1} \\
    &\le& \mu_1^2 \delta^{\mu_1,T}_{2,0} + 2\mu_1(1-\mu_1) \delta^{\mu_1,T}_{1,1} + (1-\mu_1)^2 \delta^{\mu_1,T}_{0,2} \\
    &\le& \sum_{a+b=3} {3\choose a}\mu_1^a(1-\mu_1)^b\delta^{\mu_1,T}_{a,b}\\
    &\le& \cdots\\
    &\le& \sum_{a+b=T} {T\choose a}\mu_1^a(1-\mu_1)^b\delta^{\mu_1,T}_{a,b}\\
    &=& \E[{1\over T(T+1)} (a-\mu_1 T)^+] \\
    &=&{1\over T(T+1)} \E[ (a-\mu_1 T)^+]
  \end{eqnarray*}
  
  The expectation is taken over a binomial distribution $a \sim Binomial(T,\mu_1)$.
  
  Since $\E[a] = \mu_1 T$, we have that:
  
  \begin{equation*}
\E[ (a-\mu_1 T)^+] = {1\over 2} \E[|a-\mu_1 T|] \le {1\over 2}\sqrt{\E[(a-\mu_1 T)^2]} = {1\over 2}\sqrt{T\mu_1(1-\mu_1)},
\end{equation*}which leads to the upper bound $\delta^{\mu_1,T}_{0,0} \le {\sqrt{\mu_1(1-\mu_1)}\over 2(T+1)\sqrt T}$.
  
  Thus, 
  
  \begin{eqnarray*}
    DP(\mu_1,T) &\ge& DP(\mu_1,1) - \sum_{T=1}^\infty {\sqrt{\mu_1(1-\mu_1)}\over 2(T+1)\sqrt T} \\
    &\geq& \mu_1 - {\sqrt{\mu_1(1-\mu_1)}\over 2}\sum_{T=1}^\infty {1\over T^{3/2}} \\
    &\ge& \mu_1 - {\sqrt{\mu_1(1-\mu_1)}\over 2}(1 + \int_1^\infty{1\over T^{3/2}}dT)\\
    &=& \mu_1 - {\sqrt{\mu_1(1-\mu_1)}\over 2}(1+2)\\
    &=& \mu_1 - {3\sqrt{\mu_1(1-\mu_1)}\over 2}
  \end{eqnarray*}
  When $\mu_1 \ge 0.9$, ${3\sqrt{\mu_1(1-\mu_1)}\over 2} \le 0.45$, thus $DP(\mu_1,T) \ge \mu_1 - 0.45 \ge {\mu_1\over 2}$.

\section{Proof of Theorem \ref{Theorem_LB} when $T$ is unknown}
This proof is suggested by an anonymous  reviewer of our paper during the review process. 
We thank the reviewer   for  the idea. 
\begin{proposition}\label{Proposition_LB}
  In KCMAB, if an algorithm guarantees an $o(T^\alpha)$ regret upper bound for any $T$ and any $\alpha > 0$, then the algorithm must pay $\Omega\left(\sum_{i=2}^N {\Delta_i\log T\over KL(D_i,D_1)}\right)$ for compensation. 
\end{proposition}

Choose $N_1^*(\varepsilon) = {9\over 2\Delta_2^2}\log {9\over \varepsilon\Delta_2^2}$, and $N_1^{**}(\varepsilon)$ be the time step such that with probability $1-{\varepsilon \over 2}$, $N_1(t) > {t\over 2}$ for any $t \ge N_1^{**}(\varepsilon)$. Notice that $N_1^{**}(\varepsilon)$ must exists since it does not depend on $T$ and the algorithm has $o(T)$ regret in expectation.

Now choose $T_1^*(\varepsilon) = \max \{2N_1^*(\varepsilon), N_1^{**}(\varepsilon) \}$. Note that $T_1^*(\varepsilon)$ does not depend on $T$ as well. The probability that $\exists t > T_1^*(\varepsilon)$, with $\hat{\mu}_1(t) < \mu_1 - {\Delta_2 \over 3}$ can be upper bounded by:  
\begin{eqnarray}
\nonumber&&\Pr[\exists t > T_1^*(\varepsilon), \hat{\mu}_1(t) < \mu_1 - {\Delta_2 \over 3}]\\ 
\nonumber&\le& \Pr[N_1(T_1^*(\varepsilon)) \le N_1^*(\varepsilon)] 
+ \Pr[\{N_1(T_1^*(\varepsilon)) > N_1^*(\varepsilon)\} \land \{\exists t > T_1^*(\varepsilon), \hat{\mu}_1(t) < \mu_1 - {\Delta_2 \over 3}\}]\\
\nonumber&\le& \Pr[N_1(T_1^*(\varepsilon)) \le N_1^*(\varepsilon)] +
\Pr[\exists n > N_1^*(\varepsilon) \text{ s.t. } N_1(t) = n, \hat{\mu}_1(t) < \mu_1 - {\Delta_2 \over 3}]\\
&\le& \Pr[N_1(T_1^*(\varepsilon)) \le N_1^*(\varepsilon)] +
\sum_{n > N_1^*(\varepsilon)} \left(\Pr[N_1(t) = n, \hat{\mu}_1(t) < \mu_1 - {\Delta_2 \over 3}]\right)\label{eq_333}
\end{eqnarray}

The first term in \eqref{eq_333} has upper bound ${\varepsilon \over 2}$ by definition of $N_1^{**}(\varepsilon)$ and $T_1^*(\epsilon)$.

As for the second term in \eqref{eq_333}, notice that $\{N_1(t) = n, \hat{\mu}_1(t) < \mu_1 - {\Delta_2\over 3}\}$ implies the first $n$ feedbacks of arm $1$ have an empirical mean less than $\mu_1 - {\Delta_2\over 3}$. By Chernoff Bound, $\Pr[N_1(t) = n, \hat{\mu}_1(t) < \mu_1 - {\Delta_2 \over 3}] \le \exp(-2n\Delta_2^2/9)$. Therefore $\sum_{n > N_1^*(\varepsilon)} \Pr[N_1(t) = n, \hat{\mu}_1(t) < \mu_1 - {\Delta_2 \over 3}] \le {9\over 2\Delta_2^2} \exp(-2N_1^*(\varepsilon)\Delta_2^2/9) \le {\varepsilon \over 2}$. 

This means that with probability at least $1 - \varepsilon$, for any $t > T_1^*(\varepsilon)$, $\hat{\mu}_1(t) \ge \mu_1 - {\Delta_2 \over 3}$.

Similarly, for any sub-optimal arm $i$, we can find $T_i^*(\varepsilon)$ such that with probability $1-\varepsilon$, $\hat{\mu}_i(t) \le \mu_i + {\Delta_2 \over 3}$ for any $t > T_1^*(\varepsilon)$. The only difference in this argument is that we need to use Fact \ref{Fact_Lai} instead of the fact that the algorithm has $o(T)$ regret in expectation.

Let $T^*(\varepsilon) = \max_i T_i^*(\varepsilon)$.  We know that after $T^*(\varepsilon)$, with probability at least $1-N\varepsilon$, pulling arm $i$ once needs at least $\hat{\mu}_1(t) - \hat{\mu}_i(t) \ge {\Delta_i \over 3}$ for compensation.

Before time $T^*(\varepsilon)$, every arm can be pulled for at most $T^*(\varepsilon)$ time steps. As $T$ goes to infinity, by Fact \ref{Fact_Lai}, every sub-optimal arm $i$ needs to be pulled for at least $(1-\varepsilon){\log T\over KL(D_i,D_1)}$ times. Thus the player needs to pay at least $\left((1-\varepsilon){\log T\over KL(D_i,D_1)} - T^*(\varepsilon)\right) \times {\Delta_i \over 3}$ for compensation on arm $i$ until time $T$.

Taking $T$ going to infinity and setting $\varepsilon = {1\over 2N}$, since $T^*(\varepsilon)$ does not depend on $T$, the total compensation is  
\begin{equation*}
\Omega(\sum_{i=2}^N{\Delta_i\log T\over KL(D_i,D_1)})
\end{equation*}

\section{Proof of Theorem \ref{Theorem_UCB}}

After the first $N$ time steps, every arm $i$ has $\hat{\mu}_i(t) = M_i(t)/N_i(t)$.

Notice that we always choose the arm $i$ with maximum value $\hat{\mu}_i(t) + r_i(t)$. Thus the arm $a(t)$ satisfies the following inequality:
\begin{equation*}
\hat{\mu}_{a(t)}(t) + r_{a(t)}(t) \ge \max_j {\hat{\mu}_j(t) + r_j(t)} \ge \max_j \hat{\mu}_j(t)
\end{equation*} This means that we need to pay at most $r_{a(t)}(t)$ for compensation.

For each sub-optimal arm $i \ne 1$, if it is chosen at time $t$, then we must have $\hat{\mu}_i(t) + r_i(t) \ge \hat{\mu}_1(t) + r_1(t)$. Since $\mu_1 = \mu_i + \Delta_i$, we have:

\begin{equation*}
    \hat{\mu}_i(t) + \mu_1 + 2r_i(t) \ge \mu_i+r_i(t)+\hat{\mu}_1(t) + r_1(t) + \Delta_i
\end{equation*}

This implies that one of the following three events must happen:

\begin{equation*}
  \mathcal{A}^{UCB}_i(t) = \{\hat{\mu}_i(t) \ge \mu_i+r_i(t)\}
\end{equation*}
\begin{equation*}
  \mathcal{B}^{UCB}(t) = \{\mu_1 \ge \hat{\mu}_1(t)+r_1(t)\}
\end{equation*}
\begin{equation*}
  \mathcal{C}^{UCB}_i(t) = \{2r_i(t) \ge \Delta_i\}
\end{equation*}

Thus $\E[N_i(T)] \le \sum_{t=1}^T (\Pr[\mathcal{A}^{UCB}_i(t)] + \Pr[\mathcal{B}^{UCB}(t)] + \Pr[\mathcal{C}^{UCB}_i(t)])$.

Notice that $r_i(t) = \sqrt{2\log t\over N_i(t)}$, then if $N_i(t) > {8\log T\over \Delta_i^2}$, event $\mathcal{C}^{UCB}_i(t)$ can not happen, which means $\sum_{t=1}^T \Pr[\mathcal{C}^{UCB}_i(t)] \le{8\log T\over \Delta_i^2} $. As for events $\mathcal{A}^{UCB}_i(t)$ and $\mathcal{B}^{UCB}(t)$, we have the following fact given by Chernoff-Hoeffding's inequality.

\begin{fact}\label{Fact_UCB_1}
	For any arm $i$, we have:
	\begin{equation*}
\sum_{t=1}^T \Pr[ \hat{\mu}_i(t) \ge \mu_i+r_i(t)] \le {1\over t^2}
\end{equation*}
\begin{equation*}
\sum_{t=1}^T \Pr[ \hat{\mu}_i(t) \le \mu_i-r_i(t)] \le {1\over t^2}
\end{equation*}
\end{fact}
By Fact \ref{Fact_UCB_1}, we have $\sum_{t=1}^T (\Pr[\mathcal{A}^{UCB}_i(t)] + \Pr[\mathcal{B}^{UCB}(t)]) \le {\pi^2 \over 3}$. 

If arm $i$ has been pulled for $N_i(T)$ times, we need to pay compensation for at most $\sum_{k=1}^{N_i(T)} \sqrt{2\log T \over k} \le \sqrt{8N_i(T)\log T}$, then 
\begin{equation*}
Com_i(T) \le \E_{N_i(T)}\left[\sqrt{8N_i(T)\log T}\right] \le \sqrt{8\E[N_i(T)]\log T} \le {8\log T\over \Delta_i} + {\pi^2\over 3}.
\end{equation*}



As for arm 1, we can see that when $N_1(t) = \max_i N_i(t)$ and $a(t) = 1$, we do not need to pay compensation. The reason is that $\hat{\mu}_1(t)+r_1(t)\ge \hat{\mu}_i(t)+r_i(t)$ and $r_i(t)\ge r_1(t)$ imply $\hat{\mu}_1(t)\ge \hat{\mu}_i(t)$.

Thus, let $N_1'(T) = \max_{i\ne 1} N_i(T)$, we know that we only need to pay compensation for pulling arm $1$ when $N_1(t) \le N_1'(T)$. 

Notice that we have $\E[N_1'(T)] \le \sum_{i\ne 1}\E[N_i(T)] \le \sum_{i\ne 1} {8\log T\over \Delta_i^2} + {N\pi^2\over 3}$. Thus, the compensation we need to pay on arm $1$ satisfies 

\begin{equation*}
Com_1(T) \le \E_{N_1'(T)}\left[\sqrt{8N_1'(T)\log T}\right]\le \sqrt{8\E[N_1'(T)]\log T} \le \sum_{i\ne 1} {8\log T\over \Delta_i} + {N\pi^2\over 3}.
\end{equation*}

Summing over all sub-optimal arms and the optimal arm, in Algorithm \ref{Algorithm_UCB}, we have

\begin{equation*}
Com(T) = \sum_i Com_i(T) \le \sum_{i=2}^N {16\log T\over \Delta_i} + {2N\pi^2\over 3}.
\end{equation*}

\section{Proof for Theorem \ref{Theorem_Epoch}}

  Notice that only if we choose to explore arm $j$, we need to pay the compensation. Now consider the expected compensation we need to pay on explore arm $j$ for the $k$-th time, which can be written as $\E[\max_i (\hat{\mu}_i(t_{j}^{\varepsilon}(k)) - \hat{\mu}_{j}(t_{j}^{\varepsilon}(k)))]$.

  Then we can have:

  \begin{eqnarray}
   && \E[\max_i (\hat{\mu}_i(t_{j}^{\varepsilon}(k)) - \hat{\mu}_{j}(t_{j}^{\varepsilon}(k)))]
    \nonumber\\
    &=& \E[\max_i (\hat{\mu}_i(t_{j}^{\varepsilon}(k)) - \mu_i + \mu_i - \mu_{j} + \mu_{j} - \hat{\mu}_{j}(t_{j}^{\varepsilon}(k)))] \\
    \nonumber&\le&\E[\max_i (\hat{\mu}_i(t_{j}^{\varepsilon}(k)) - \mu_i)] + \E[\max_i (\mu_i - \mu_{j})] + \E[(\mu_{j} - \hat{\mu}_{j}(t_{j}^{\varepsilon}(k)))]  \\
    \label{eq_1}&=& \E[\max_i (\hat{\mu}_i(t_{j}^{\varepsilon}(k)) - \mu_i)] + \Delta_{j} + \E[(\mu_{j} - \hat{\mu}_{j}(t_{j}^{\varepsilon}(k)))]\\
    \label{eq_2}&=& \E[\max_i (\hat{\mu}_i(t_{j}^{\varepsilon}(k)) - \mu_i)] + \Delta_{j} 
  \end{eqnarray}

  Eq. \eqref{eq_1} is because that $\E[\max_i (\mu_i - \mu_{j})] = \max_i (\mu_i - \mu_{j}) = \mu_1 - \mu_{j} = \Delta_{j}$, and Eq. \eqref{eq_2} is because that whether we choose to explore arm $j$ are independent with the its observations.

  Now we consider the value $\E[\max_i (\hat{\mu}_i(t) - \mu_i)]$. It is upper bounded by $\sum_i \E[(\hat{\mu}_i(t) - \mu_i)^+]$. When $j$ is explored for $k$ times, we know every arm must have be chosen for at least $k$ times. Notice that these feedbacks are independent with whether we choose to explore as well. Thus, we have:

  \begin{eqnarray}
    \nonumber\sum_i \E[(\hat{\mu}_i(t_{j}^{\varepsilon}(k)) - \mu_i)^+] &=& {1\over 2} \sum_i \E[|\hat{\mu}_i(t_{j}^{\varepsilon}(k)) - \mu_i|]\\
    \nonumber&\le& {1\over 2} \sum_i \sqrt{\E[(\hat{\mu}_i(t_{j}^{\varepsilon}(k)) - \mu_i)^2]}\\
    \nonumber&\le& {1\over 2} \sum_i \sqrt{1\over 4k}\\
    \nonumber&=& {N\over 4\sqrt k}
  \end{eqnarray}

  Suppose arm $j$ has been exploration for $n_j(T)$ times until time $T$. 
Then 
  
  \begin{equation*}
\sum_{k=1}^{n_j(T)} \E[\max_i (\hat{\mu}_i(t_{j}^{\varepsilon}(k)) - \hat{\mu}_{j}(t_{j}^{\varepsilon}(k)))] \le \sum_{k=1}^{n_j(T)} \left({N\over 4\sqrt{k}} + \Delta_j\right) \le n_j(T)\Delta_j + {N\over 2}\sqrt{n_j(T)}
\end{equation*}

  Notice that when $\epsilon = {cN\over \Delta_2^2}$, $\E[n_j(T)] = {c\log T\over \Delta_2^2}$, then $\E[n_j(T)\Delta_j + {N\over 2}\sqrt{n_j(T)}] \le {c\Delta_j\log T\over \Delta_2^2} + {N\over 2\Delta_2}\sqrt{c\log T}$. Thus the total compensation is upper bounded by:

  \begin{equation*}
    \sum_{i=2}^N {c\Delta_i\log T\over \Delta_2^2} + {N^2\over 2\Delta_2}\sqrt{c\log T}
  \end{equation*}
  
\section{Proof for Theorem \ref{Theorem_TS}}

We first  give four important lemmas, which come from the analysis of TS policy in previous works \citep{agrawal2013further,Kaufmann2012Thompson}. Their proofs can be modified slightly to work for our Algorithm \ref{Algorithm_TS}. 

\begin{restatable}{lemma}{LemmaTSOne}\label{Lemma_TS_1}(Theorem 1 in \cite{agrawal2013further})
	In Algorithm \ref{Algorithm_TS}, summing over all possible rounds $(t,t+1)$, we have that for all $i\ne 1$ and $\varepsilon < \Delta_i$:
	\begin{equation*}
	\sum_{(t,t+1)} \Pr[a_2(t+1) = i] = {2\over (\Delta_i-\varepsilon)^2} \log T + O\left({1\over \varepsilon^4}\right)
	\end{equation*}
\end{restatable}

\begin{lemma}\label{Lemma_TS_2}(Proposition 1 in \cite{Kaufmann2012Thompson})
	In Algorithm \ref{Algorithm_TS}, summing over all possible rounds $(t,t+1)$, we have that 
	\begin{equation*}
	\sum_{(t,t+1)} \Pr[N_1(t) \le t^b] \le C(\bm{\mu})
	\end{equation*}
	holds for some constant $b = b(\bm{\mu})\in (0,1)$. 
\end{lemma}

\begin{lemma}\label{Lemma_TS_3}(Lemma 2 in \cite{agrawal2013further})
	In Algorithm \ref{Algorithm_TS}, summing over all possible rounds $(t,t+1)$, for any $i\ne 1$, we have that 
	\begin{equation*}
	\sum_{(t,t+1)} \Pr[a_1(t) = i, \hat{\mu}_i(t) \ge \mu_i + {\Delta_i\over 2}] \le {4\over \Delta_i^2} + 1
	\end{equation*}
\end{lemma}

\begin{restatable}{lemma}{LemmaTSFour}\label{Lemma_TS_4}
	In Algorithm \ref{Algorithm_TS}, 
	\begin{equation*}
	\forall i, \Pr[|\theta_i(t) - \hat{\mu}_i(t)| \ge r_i(t)] \le {1\over t^2}
	\end{equation*}
	where $r_i(t) = \sqrt{2\log t\over N_i(t)}$.
\end{restatable}

We do not provide the proofs of Lemma \ref{Lemma_TS_1}, Lemma \ref{Lemma_TS_2} and Lemma \ref{Lemma_TS_3} since their proofs are almost the same as in \cite{agrawal2013further} and \cite{Kaufmann2012Thompson}. As for Lemma  \ref{Lemma_TS_4}, although the proof is similar, the statement is not the same. We provide its proof in the end of this section. 



Firstly, we analyze the regret bound of Algorithm \ref{Algorithm_TS}.





Lemma \ref{Lemma_TS_1} shows that the regret during sample steps are bounded, now we come to the regret during empirical steps.

We use the following four events to help our analysis:

\begin{equation*}
  \mathcal{A}^{TS}_i(t) = \{\hat{\mu}_i(t) \ge \mu_i + {\Delta_i\over 2}\}
\end{equation*}
\begin{equation*}
  \mathcal{B}^{TS}(t) = \{\hat{\mu}_1(t) + r_1(t) \le \mu_1\}
\end{equation*}
\begin{equation*}
  \mathcal{C}^{TS}_i(t) = \{2r_1(t) \ge \Delta_i\}
\end{equation*}
\begin{equation*}
  \mathcal{D}^{TS}(t) = \{N_1(t) > t^b\}
\end{equation*}

Then:
\begin{eqnarray*}
\sum_{(t,t+1)} \Pr[a_1(t) = i] &\le& \sum_{(t,t+1)} \Pr[ \mathcal{A}^{TS}_i(t)\cap \{a_1(t) = i\}] + \sum_{(t,t+1)}\Pr[\mathcal{B}^{TS}(t)\cap \{a_1(t) = i\}]\\
&& +  \sum_{(t,t+1)}\Pr[ \mathcal{C}^{TS}_i(t) \cap \mathcal{D}^{TS}(t)\cap \{a_1(t) = i\}] + \sum_{(t,t+1)} \Pr[ \neg \mathcal{D}^{TS}(t)\cap \{a_1(t) = i\}]\\
&& +  \sum_{(t,t+1)}\Pr[ \neg \mathcal{A}^{TS}_i(t) \cap \neg \mathcal{B}^{TS}(t) \cap \neg \mathcal{C}^{TS}_i(t)\cap \{a_1(t) = i\}])
\end{eqnarray*}


Lemma \ref{Lemma_TS_3} shows that $\sum_{(t,t+1)} \Pr[\mathcal{A}^{TS}_i(t)\cap \{a_1(t) = i\}] \le {4\over \Delta_i^2} + 1$.
Using Fact \ref{Fact_UCB_1}, $\sum_{(t,t+1)}\Pr[\mathcal{B}^{TS}(t)\cap \{a_1(t) = i\}] \le \sum_{(t,t+1)}\Pr[\mathcal{B}^{TS}(t)] \le {\pi^2 \over 6}$.

Then we consider $t$ such that $ \mathcal{C}^{TS}_i(t) \cap \mathcal{D}^{TS}(t)$ happens. By definition, we can see that ${\Delta_i\over 2} \le r_1(t) = \sqrt{2\log t\over N_1(t)} \le \sqrt{2\log t\over t^b}$. Thus there exists $t_i = f(i,\bm{\mu})$ such that for all $t \ge t_i$, $\Pr[\mathcal{C}^{TS}_i(t) \cap \mathcal{D}^{TS}(t)] = 0$. This implies that: 
\begin{equation*}
\sum_{(t,t+1)}\Pr[ \mathcal{C}^{TS}_i(t) \cap \mathcal{D}^{TS}(t)\cap \{a_1(t) = i\}] \le \sum_{(t,t+1)} \Pr[ \mathcal{C}^{TS}_i(t) \cap \mathcal{D}^{TS}(t)] \le t_i
\end{equation*}

$\sum_{(t,t+1)} \Pr[ \neg \mathcal{D}^{TS}(t)\cap \{a_1(t) = i\}]$ is upper bounded by Lemma \ref{Lemma_TS_2}, which is $C(\bm{\mu})$.

$\neg \mathcal{A}^{TS}_i(t) \cap \neg \mathcal{B}^{TS}(t) \cap \neg \mathcal{C}^{TS}_i(t)\cap \{a_1(t) = i\}$ cannot happen since under the first three events we have:
\begin{equation*}
\hat{\mu}_1(t) > \mu_1 - r_1(t) > \mu_1 - {\Delta_i\over 2} = \mu_i + {\Delta_i\over 2} > \hat{\mu}_i(t),
\end{equation*}
which contradict with $\{a_1(t) = i\}$.

Thus, we have that $\sum_{i=2}^N \left(\sum_{(t,t+1)} \Pr[a_1(t) = i] \right) \le N\left(1+{\pi^2\over 6}\right) + \sum_{i=2}^N \left({4\over \Delta_i^2} +t_i\right) + C(\bm{\mu}) = F_1(\bm{\mu})$ for some function $F_1$, and it is independent with time horizon $T$.





Adding the regret during sample steps, the total regret of Algorithm \ref{Algorithm_TS} is upper bounded by $\sum_i {2\Delta_i\over (\Delta_i - \varepsilon)^2} \log T + O\left({1\over \varepsilon^4}\right) + F_1(\bm{\mu})$.

Now we consider the compensation. Notice that in empirical steps we always choose the arm with the largest empirical mean, thus we do not need to pay any compensation in this time slot. Because of this, we can focus on the compensation in sample steps. To do so, we define a event $\mathcal{E}^{TS}(t)$ as following:

\begin{equation*}
  \mathcal{E}^{TS}(t) = \{\forall i, |\theta_i(t) - \hat{\mu}_i(t)| \le r_i(t)\}
\end{equation*}

Then the total compensation can be written as

\begin{equation}\label{eq_m120}
\E[\sum_{(t,t+1): t+1 < T} c_i(t+1)] \le \E[\sum_{(t,t+1): t+1 < T} \I[\mathcal{E}^{TS}(t)]c_i(t+1)]  + \E[\sum_{(t,t+1): t+1 < T} \I[\neg \mathcal{E}^{TS}(t)] c_i(t+1)]
\end{equation}

Lemma \ref{Lemma_TS_4} shows that $\E[\sum_{(t,t+1): t+1 < T} \I[\neg \mathcal{E}^{TS}(t)]] \le {N\pi^2\over 6}$, thus the second term in Eq. \eqref{eq_m120} has upper bound $ {N\pi^2\over 6}$ as well.

Now we consider the first term in Eq. \eqref{eq_m120}. Here the compensation we need to pay is $c_i(t+1) = \max_j \hat{\mu}_j(t+1) - \hat{\mu}_{a_2(t+1)}(t+1)$. 


If $a_1(t) = a_2(t+1) = i$ and $\hat{\mu}_{i}(t+1) \ge \hat{\mu}_i(t)$, we have $c_i(t+1) = 0 < {1\over N_i(t)}$.

If $a_1(t) = a_2(t+1) = i$ but $\hat{\mu}_{i}(t+1) < \hat{\mu}_i(t)$, then we know that $\max_j \hat{\mu}_j(t+1) \le \hat{\mu}_i(t)$, thus $c_i(t+1) \le  \hat{\mu}_i(t) -  \hat{\mu}_i(t+1) \le {1\over N_i(t)}$.

If $a_1(t) =k, a_2(t+1) = i$ and $k\ne i$, then $c_i(t+1) = \max_j \hat{\mu}_j(t+1) - \hat{\mu}_{i}(t+1) \le \hat{\mu}_k(t) + {1\over N_k(t)} - \hat{\mu}_i(t)$.

Thus, if $a_1(t) = a_2(t+1)$, we need to pay at most  ${1\over N_i(t)}$ for compensation. Otherwise, we need to pay at most 
$ (\hat{\mu}_{a_1(t)}(t)  - \hat{\mu}_{a_2(t+1)}(t)) + {1\over N_{a_1(t)}(t)} $.


Notice that under event $\mathcal{E}(t)$, $\hat{\mu}_{a_2(t+1)}(t) + r_{a_2(t+1)}(t) \ge \theta_{a_2(t+1)}(t) \ge \theta_{a_1(t)}(t) \ge \hat{\mu}_{a_1(t)}(t) - r_{a_1(t)}(t)$. Thus if $a_1(t) \ne a_2(t+1)$, $c_i(t+1) \le r_{a_1(t)}(t) + r_{a_2(t+1)}(t) + {1\over N_{a_1(t)}(t)}$. Then we can treat the total compensation as following: we first pay $r_{a_1(t}(t) + {1\over N_{a_1(t)}(t)}$ on arm $a_1(t)$ in the empirical step, and then pay $r_{a_2(t+1)}(t)$ on arm $a_2(t+1)$ in the sample step. By this method, we can upper bound the compensation we need to pay on pulling sub-optimal arm $i$ as $\sum_{\tau=1}^{N_i(T)} \sqrt{2\log T\over \tau} + {1\over \tau} \le \log T + \sqrt{8N_i(T)\log T}$ under the event $\mathcal{E}^{TS}(t)$.

By regret analysis, $\E[N_i(T)] \le {2\over (\Delta_i-\varepsilon)^2} \log T + O\left({1\over \varepsilon^4}\right) + t_i + {4\over \Delta_i^2} + {\pi^2+6\over 6} + C(\bm{\mu})$, thus the compensation $Com_i(T)$ is upper bounded by $ {4\over \Delta_i - \varepsilon} \log T + O\left({1\over \varepsilon^4}\right) + t_i + {4\over \Delta_i^2} + {\pi^2+6\over 6} + C(\bm{\mu}) + \log T$.

As for arm 1, when $a_1(t) = a_2(t+1) = 1$, we only need to pay ${1\over N_1(t)}$ for compensation.  The expected number of time steps that we do not have $a_1(t) = a_2(t) = 1$ is at most $\sum_{i=2}^N {2\over (\Delta_i - \varepsilon)^2} \log T + F(\bm{\mu}) + O\left({N\over \varepsilon^4}\right)$, which is given by the regret analysis. This means the compensation on pulling arm 1 is upper bounded by $\sum_{i=2}^N  {4\over \Delta_i - \varepsilon} \log T + F_1(\bm{\mu})+ O\left({N\over \varepsilon^4}\right) + \log T$.

Thus, the first term in Eq. \eqref{eq_m120} has upper bound $\sum_{i=2}^N {8\over \Delta_i - \varepsilon} \log T + F_1(\bm{\mu})+ O\left({N\over \varepsilon^4}\right) + N\log T + \sum_{i=2}^N \left(t_i + {4\over \Delta_i^2} + {\pi^2+6\over 6} + C(\bm{\mu})\right)$. After adding the upper bound ${N\pi^2\over 6}$ of the second term and setting $F_2(\bm{\mu}) = F_1(\bm{\mu}) +{N\pi^2\over 6}+\sum_{i=2}^N \left(t_i + {4\over \Delta_i^2} + {\pi^2+6\over 6} + C(\bm{\mu})\right) $, we have that \begin{equation*}
	Com(T) \le \sum_i {8 \over \Delta_i - \varepsilon} \log T + N\log T + O\left({N\over \varepsilon^4}\right) + F_2(\bm{\mu}).
\end{equation*}

\subsection{Proof of Lemma \ref{Lemma_TS_4}}

Since $\theta_i(t)$ only depends on the values of $(\alpha_i(t), \beta_i(t))$ but is independent of the random history, we can fix the pair $(\alpha_i(t), \beta_i(t))$ to prove the inequality, and then the inequalities hold also for  $(\alpha_i(t), \beta_i(t))$ as random variables.
	
	\begin{eqnarray}
		\nonumber\Pr[\theta_i(t) > \hat{\mu}_i(t) + r_i(t)] &=& 1 - F^{Beta}_{\alpha_i(t),\beta_i(t)(\hat{\mu}_i(t) + r_i(t))}\\
		\label{eq_m100}&=& 1 - (1 - F^B_{\alpha_i(t)+\beta_i(t)-1, \hat{\mu}_i(t) + r_i(t)}(\alpha_i(t) - 1))\\
		\nonumber&=& F^B_{\alpha_i(t)+\beta_i(t)-1, \hat{\mu}_i(t) + r_i(t)}(\alpha_i(t) - 1)\\
		\nonumber&\le& F^B_{\alpha_i(t)+\beta_i(t)-1, \hat{\mu}_i(t) + r_i(t)}(\hat{\mu}_i(t)(\alpha_i(t)+\beta_i(t)-1))\\
		\label{eq_m101}&\le&\exp(-(\alpha_i(t)+\beta_i(t)-1)KL(\hat{\mu}_i(t), \hat{\mu}_i(t) + r_i(t)))\\
		\label{eq_m102}&\le& \exp(-{N_i(t)r_i(t)^2\over 2})\\
		\nonumber&\le& {1\over t^2}
	\end{eqnarray}
	
Eq. \eqref{eq_m100} is given by the following Beta-Binomial Trick (Fact \ref{Fact33}), Eq. \eqref{eq_m101} is given by Chernoff-Hoeffding Inequality, and Eq. \eqref{eq_m102} follows the fact that $KL(x,y) \ge {|x-y|^2\over 2}$.

\begin{fact}\label{Fact33}(Beta-Binomial Trick)
	Let $F^{Beta}_{a,b}(x)$ be the cdf of Beta distribution with parameters $(a,b)$, let $F^B_{n,p}(x)$ be the cdf of Binomial distribution with parameters $(n,p)$. Then for any positive integers $(a,b)$, we have
	\begin{equation*}
		F^{Beta}_{a,b}(x) = 1 - F^B_{a+b-1,x}(a-1)
	\end{equation*}
\end{fact}

\end{document}